\documentclass[10pt,twocolumn]{article} 
\usepackage{simpleConference}
\usepackage{times}
\usepackage{graphicx}
\usepackage{url,hyperref}
\usepackage{cite}
\usepackage{amsmath,amssymb,amsfonts,array}
\usepackage{algorithmic}
\usepackage{textcomp}
\usepackage{xcolor}
\usepackage[latin1]{inputenc}
\usepackage{multirow}
\usepackage{booktabs}

\begin{document}

\title{Ultra-low-power Range Error Mitigation for Ultra-wideband Precise Localization}

\author{Simone Angarano\textsuperscript{1,2} \and 
        Francesco Salvetti\textsuperscript{1,2,3} \and
        Vittorio Mazzia\textsuperscript{1,2,3} \and
        Giovanni Fantin\textsuperscript{1,2} \and 
        Dario Gandini\textsuperscript{1,2} \and 
        Marcello Chiaberge\textsuperscript{1,2} \and
        \textsuperscript{1}Department of Electronics and Telecommunications, Politecnico di Torino, Turin, Italy \\
        \textsuperscript{2}PIC4SeR, Politecnico di Torino Interdepartmental Centre for Service Robotics, Turin, Italy \\
        \textsuperscript{3}SmartData@PoliTo, Big Data and Data Science Laboratory, Turin, Italy \\
        \tt{\{name.surname\}@polito.it}}

\maketitle
\thispagestyle{empty}

\begin{abstract}
Precise and accurate localization in outdoor and indoor environments is a challenging problem that currently constitutes a significant limitation for several practical applications. Ultra-wideband (UWB) localization technology represents a valuable low-cost solution to the problem. However, non-line-of-sight (NLOS) conditions and complexity of the specific radio environment can easily introduce a positive bias in the ranging measurement, resulting in highly inaccurate and unsatisfactory position estimation. In the light of this, we leverage the latest advancement in deep neural network optimization techniques and their implementation on ultra-low-power microcontrollers to introduce an effective range error mitigation solution that provides corrections in either NLOS or LOS conditions with a few mW of power. Our extensive experimentation endorses the advantages and improvements of our low-cost and power-efficient methodology.
\end{abstract}

\section{Introduction}
As Global Navigation Satellite System (GNSS) is the benchmark solution for outdoor positioning, Ultra-wideband (UWB) real-time locating systems (RTLS) have recently become the state of the art technology for localization in indoor environments \cite{tiwari2020design}. Indeed, with its high signal frequency and very narrow pulses, UWB outperforms all other wireless positioning systems like WiFi and BLE thanks to its decimeter level of precision and higher resilience to multipath effects \cite{schmid2019accuracy}.

Nevertheless, in a real-world scenario, the complexity of the environment often leads to partial or total obstruction of the signal between the transmitter and the receiver, thus causing a substantial degradation of the positioning performances. The non-line-of-sight (NLOS) condition affects the time-of-arrival (ToA) measurement introducing a positive bias in the ranging estimation \cite{otim2019effects}. Moreover, multipath components also strongly influence range estimates, especially in indoor environments where walls and furniture are often made of reflecting materials. 

\begin{figure}[t]
    \centering
    \includegraphics[width=0.9\columnwidth]{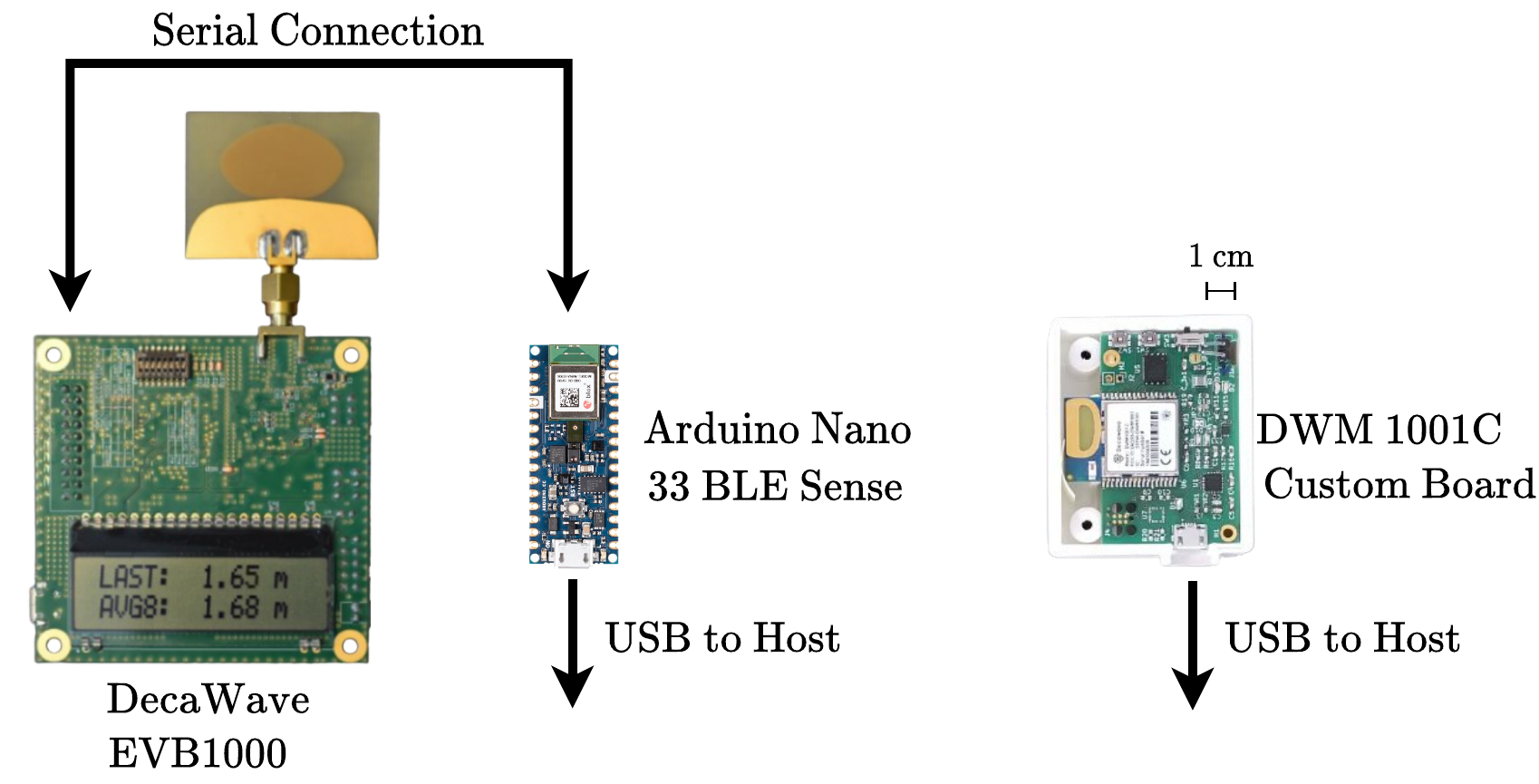}
    \small\caption{Hardware setup for ultra-low-power UWB range error mitigation. The DecaWave EVB1000 board is connected to a power supply and an external microprocessor (Arduino Nano 33 BLE Sense) that locally runs a highly optimized and power-efficient deep neural network for range error mitigation. We also show our custom board designed for the DWM1001C module for size comparison. Future works will fully integrate our methodology on our custom board, providing a compact solution for precise localization.}
    \label{fig:setup}
\end{figure}

Therefore, to achieve better localization accuracy, a mitigation algorithm is needed, and this must be robust and general enough to be effective in a large set of different scenarios. Most NLOS identification and mitigation methodologies proposed in the literature are based on channel impulse response (CIR) statistics \cite{barral2019nlos}, likelihood ratio or binary hypothesis tests \cite{silva2016ir}, and machine learning techniques. As regards the latter, several techniques have been investigated, such as representation learning models \cite{stahlke2020nlos}, support vector machines (SVM), \cite{ying2012classification} and Gaussian processes (GP) \cite{xiao2014non}. Despite the chosen methodology, the resulting mitigation algorithm must require a low computational effort to be usable in a real-world use case and consequently come out of a pure research scenario. Indeed, most applications, like robotic indoor navigation and person or object tracking, typically make use of single-board computers with limited computational capabilities and stringent power consumption requirements. 

In this research project, we propose a highly optimized deep learning model for range error mitigation that requires a few mW to compensate NLOS and LOS signals. The proposed methodology can run at high frequency on ultra-low-power microcontrollers, enabling the design of small and low-power devices for precise indoor localization. The complete setup used in our experimentation is presented in Figure \ref{fig:setup}. The main contributions of this paper are the following.

\begin{itemize}
    \item Introduce UWB range error mitigation for ultra-low-power microcontrollers with deep learning at the edge. 
    \item Modify and highly optimize a deep learning model, leveraging the latest weight quantization and graph optimization techniques for power and latency reduction.
    \item Evaluate the real-time performance of the resulting network, measuring latency, power, and energy usage with different CIR sizes.
\end{itemize}

The rest of the paper is organized as follows. Section \ref{methodology} presents the proposed methodology with a detailed explanation of the network and the adopted optimization techniques for power consumption and latency reduction. Section \ref{experiments_results} presents the experimental results and discussion after briefly describing the DeepUWB dataset used for the tests. Finally, section \ref{conclusions} summarizes the main achievements of the work and proposes future research developments.


\section{Methodology} \label{methodology}
In this section, we present the proposed methodology for the embedded implementation of a UWB error mitigation algorithm on an ultra-low-power microcontroller. We present details on the deep neural network design and the techniques adopted to optimize it, quantize all its parameters to 8 bits integers, and deploy the final model on the target board.

We model the mitigation process as presented by Angarano et al. \cite{angarano2021robust}:
\begin{equation}
    \hat{d} = d + \Delta d
\end{equation}
where the goal is to predict an estimate of the error $\Delta d$ on the UWB range measurement in order to compensate the observed quantity $\hat{d}$ and recover the true distance $d$.
Therefore, we adopt a DNN model that predicts an estimate
$\hat{y}$ of the true latent error $y = \Delta d$ as a non-linear function of the input CIR vector $X$, measured by the UWB sensor.

We denote with $K$ the number of temporal samples of the CIR. We develop our methodology so that the dimension $K$ can be changed to analyze its effects on the computational efficiency and the accuracy of the algorithm.

\subsection{Network Design}
The original design of the Range Error Mitigation Network (REMNet) presented in \cite{angarano2021robust} is adapted to be executed in real-time on a low-power microcontroller. The neural network should be fully quantized to perform all the operations with 8-bit integers and meet the real-time constraints. So, in order to overcome software and hardware limitations of standard low-power microcontroller solutions, we modify the original REMNet architecture, removing all self-attention blocks that boost the accuracy performance but compromise compatibility.

\begin{figure}[ht]
    \centering
    \includegraphics[width=0.95\columnwidth]{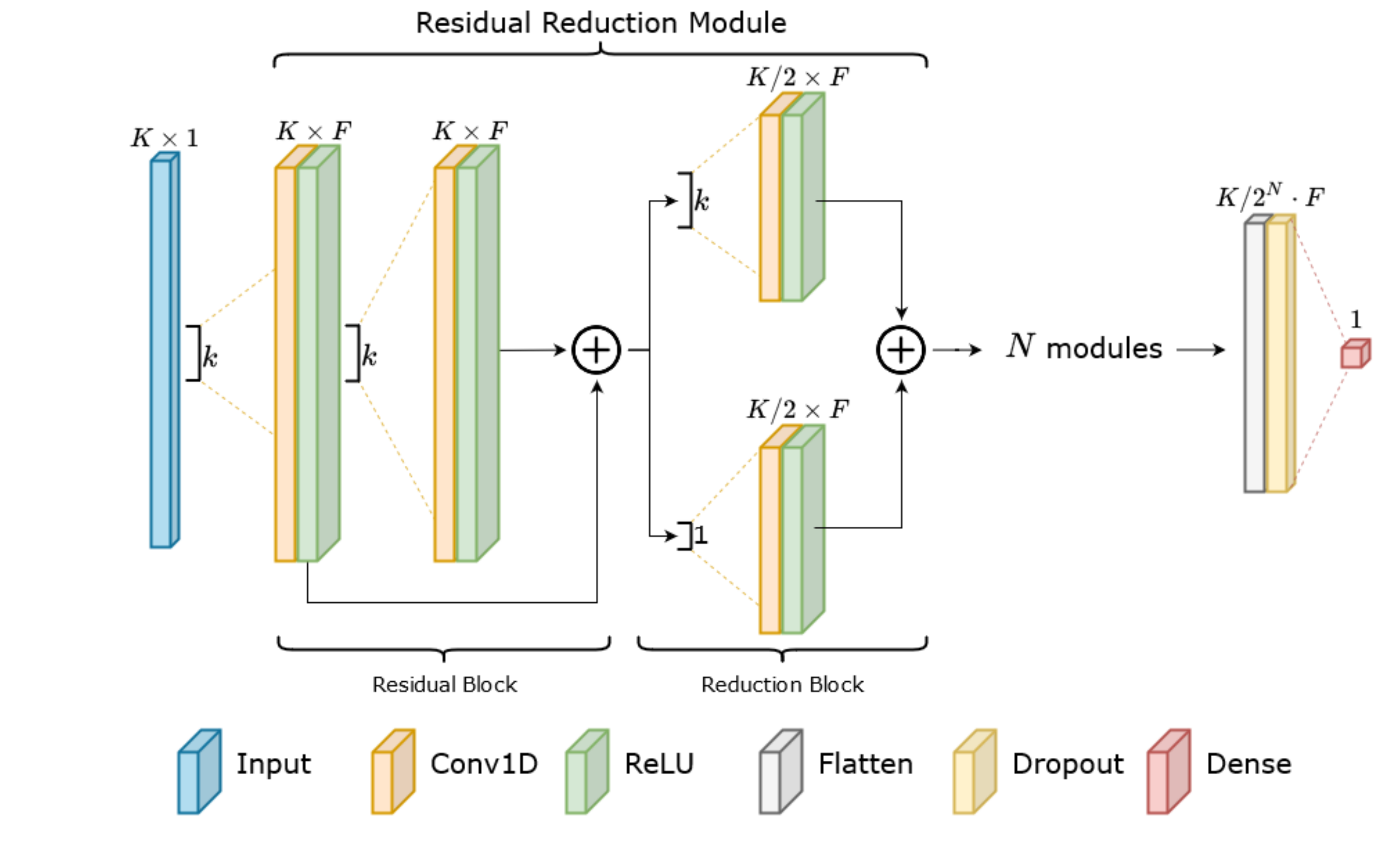}
    \small\caption{Range Error Mitigation Network (REMNet) architecture modified to ensure compatibility with the target embedded board. The input of the model is the $K\times 1$ tensor representing the CIR of the measurement.
    Subsequent $N$ Residual Reduction Modules progressively reduce the original dimension $K$. Finally, a fully connected layer composes the high-level extracted features of dimension $K/2^{N}\cdot F$ and outputs the range error estimation.}
    \label{fig:REMNet}
\end{figure}

The overall modified architecture of the REMNet model is shown in Figure \ref{fig:REMNet}. Starting from the input tensor CIR $\textbf{\textit{X}}$ of size $K \times 1$, we extract low-level features with a first 1D convolution operation with a kernel of dimension $k_0$. The core of REMNet is the residual reduction module (RRM). Firstly, the residual is computed with respect to a 1D convolution of kernel $k_n$; then, a reduction block decreases the temporal dimension $K$ with a 1D convolution with a stride of 2. The reduction block again has a residual connection characterized by a 1D convolution with a kernel $k$ of dimension 1 and stride 2 to match the temporal dimension.

Overall, each RRM block computes the following non-linear mapping function:
\begin{equation}
    \text{RRM}(\textbf{\textit{X}})  = \text{Red}(\text{Conv1D}(\textbf{\textit{X}}) + \textbf{\textit{X}})
\end{equation}
where
\begin{equation}
    \text{Red}(\textbf{\textit{X}})  = \text{Conv1D}_{s=2}(\textbf{\textit{X}}) + \text{Conv1D}_{k=1;s=2}(\textbf{\textit{X}})
\end{equation}

The network is characterized by a stack of $N$ RRM, all with ReLU as non-linear activation functions \cite{nair2010rectified}. After $N$ RRM blocks, we obtain a tensor with shape $K/2^N \times F$. We perform a flattening operation to feed a regression head composed of dropout and a fully connected layer that predicts the final estimate of the compensation value $\Delta d$. We denote with $F$ the number of features of each convolutional operation. We always use zero padding and the same value $k_n$ for each convolutional kernel.

\subsection{Network Optimization and Quantization Techniques} 

To achieve the goal of a real-time implementation, the range error mitigation technique must respect constraints on memory, power, and onboard latency. We study different graph optimization and quantization methods to reduce computational cost without compromising performance. Several techniques have been developed to increase model efficiency in the past few years\cite{jacob2018quantization}, from which the following methods are chosen. First, network pruning and layer fusing are applied to remove nodes and operations that give almost no contribution to the output. Moreover, the number of bits used to represent network parameters and activation functions is reduced by quantizing the float32 values to int8 ones. Combining these strategies strongly increases efficiency with minimal impact on performance. 

\begin{figure}[t]
    \centering
    \includegraphics[width=0.8\columnwidth]{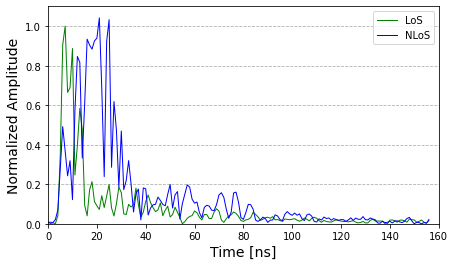}
    \small\caption{LoS and NLoS samples from the DeepUWB dataset, with normalized amplitude. In the NLoS case, the signal travels along many routes until it reaches the antenna. That makes the ToA estimation ambiguous, introducing a positive bias in the ranging measurement.}
    \label{fig:NLOS}
\end{figure}

Graph optimization is first applied to the model trained in plain float32 without quantization to investigate its effects on accuracy and dimension. Finally, a third version of the network is obtained by quantizing weights, activations, and math operations through scale and zero-point parameters.
We follow the methodology presented by Jacob et al.\cite{jacob2018quantization}, in which each weight and activation are quantized with the following equation:
\begin{equation}
    r = S(q - Z)
\end{equation}
where $r$ is the original floating-point value, $q$ the quantized integer value, and $S$ and $Q$ are the quantization parameters (respectively scale and zero point).
A fixed-point multiplication approach is adopted to cope with the non-integer scale of $S$. This strategy drastically reduces memory and computational demands due to the high efficiency of integer computations on microcontrollers.
The final step is to convert and import the quantized model into the embedded application system. As most microcontrollers do not have the resources to run a filesystem, we provide the network in a C source file that can be included in the program binary and loaded directly into memory, as suggested by \cite{tinyml}.
All the results obtained with the models at different quantization steps are presented in Section \ref{experiments_results}.


\section{Experiments and Results} \label{experiments_results} 
In this section, we perform an experimental evaluation of different optimized versions of REMNet. Moreover, we test the accuracy and performance of the network on a low-cost microcontroller-based development board, reporting inference speed, and power consumption.

\subsection{The DeepUWB Dataset}

\begin{figure}[t]
    \centering
    \includegraphics[width=0.9\columnwidth]{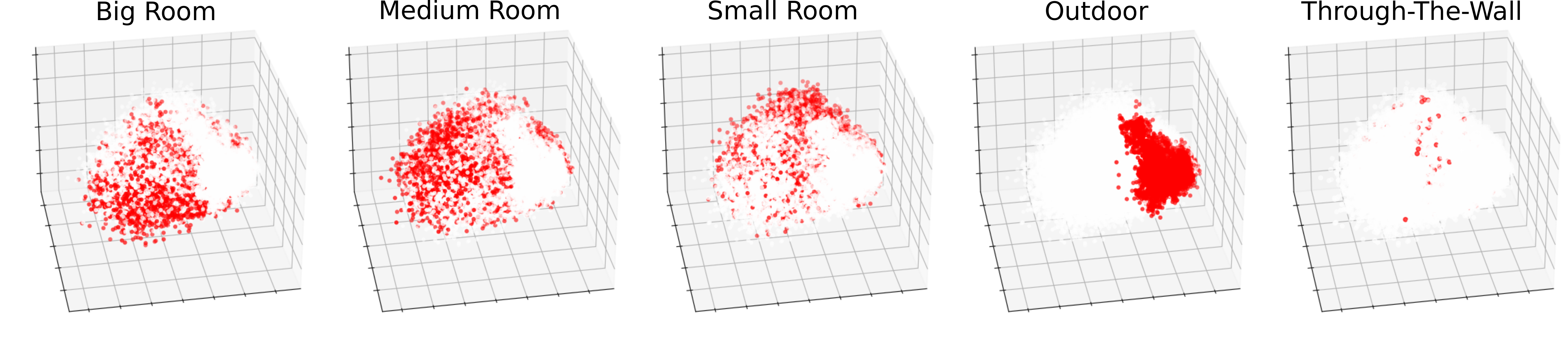}
    \includegraphics[width=0.9\columnwidth]{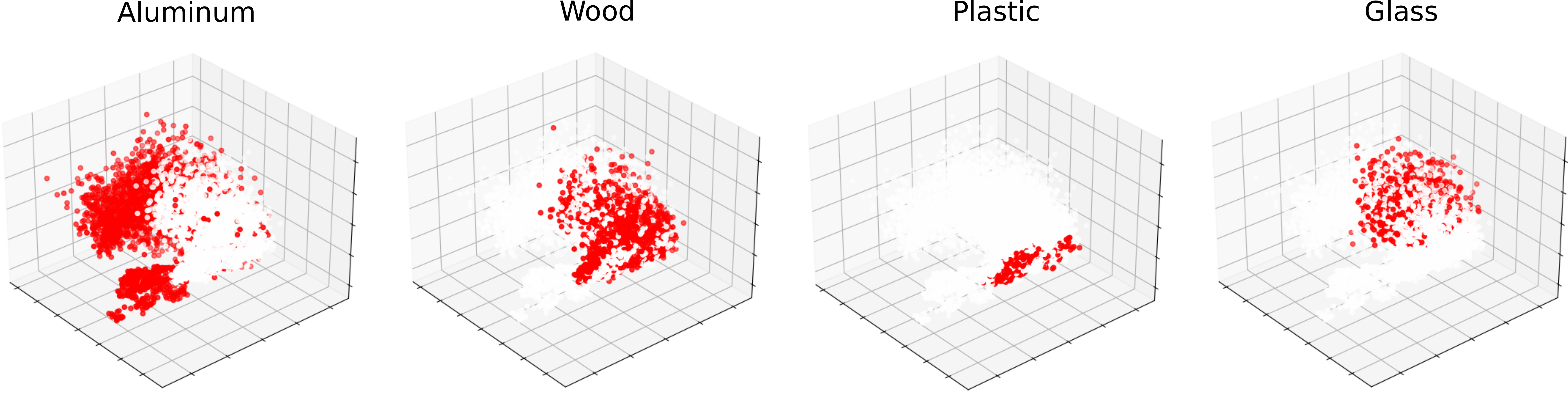}
    \small\caption{Principal Component Analysis representation of DeepUWB benchmark dataset, showing different spatial configurations for different rooms and materials. The original CIR dimensions are projected into a three-dimensional space. The first row shows the data point projection divided into the five considered environments. On the other hand, the second row highlights the effects of materials on signal propagation. It is clear how different molecular structures affect the signal in different ways.}
    \label{fig:PCA}
\end{figure}

In the following experiments, we employ the indoor samples of the DeepUWB dataset presented in \cite{angarano2021robust} and publicly available on Zenodo\footnote{http://doi.org/10.5281/zenodo.4290069}.
The data is obtained using DecaWave EVB1000 transmitters and taking several LOS and NLOS in different indoor and outdoor environments in the presence of various types of obstacles. Figure \ref{fig:NLOS} presents a comparison between LoS and NLoS samples from the dataset. Range estimates taken in NLoS conditions are typically positively biased \cite{otim2019effects}.

For each of the 55,000 measures, both ground-truth distance and the one given by the UWB boards are included, as well as the environment scenario, the obstacle materials, and the CIR vector used as input for REMNet. Three differently sized rooms are selected for indoor measurements to cover various office-like situations: a large one (10m x 5m), a medium one (5m x 5m), and a small one (5m x 3.5m). Regarding obstacles, various typical objects for an indoor scenario are used to cover a wide range of materials, including plastic, glass, metal, and wood.

Figure \ref{fig:PCA} shows the result of Principal Component Analysis (PCA) on DeepUWB: it is noticeable that the three indoor scenarios occupy close areas in the 3D space, very distinct from outdoor and through-the-wall measurements. That is due to the presence of strong multipath components. Measurements taken in the presence of different materials tend to occupy different regions, with heavier and more screening ones, such as aluminum, being highly concentrated and distant from materials like plastic and wood.

\subsection{Experimental Setting} \label{experimental_setting}
We keep aside the medium-sized room measurements as our primary scope is to evaluate the effectiveness of REMNet in compensating the error for general indoor scenarios. In total, 36023 and 13210 training and testing data points are used, respectively. LoS and NLoS samples are kept together in the sets to evaluate the performance of the network in both cases. 
Real-time range mitigation with the whole CIR vector could be very computationally intensive \cite{zeng2018nlos}. For this reason, a study is conducted on the number of samples necessary to have an acceptable error correction. Then, the Tensorflow Lite \footnote{https://www.tensorflow.org/lite}
framework is used to perform graph optimization and to quantize weights, activations, and math operations. The final test measures the inference frequency of the model deployed on an Arduino Nano 33 BLE Sense\footnote{https://store.arduino.cc/arduino-nano-33-ble-sense}, alongside its power usage. 

\begin{figure}[t]
    \centering
    \includegraphics[width=0.8\columnwidth]{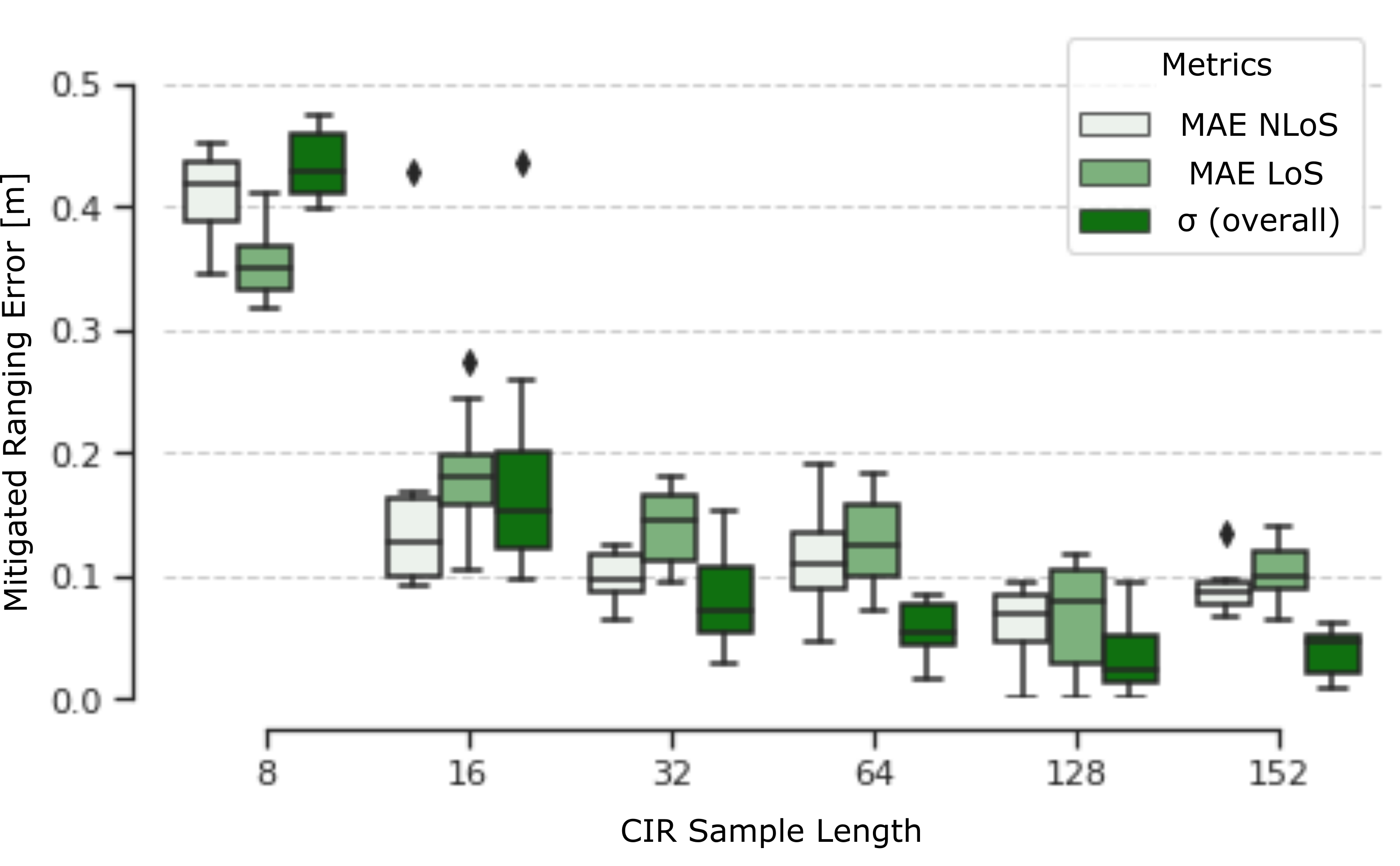}
    \small\caption{REMNet performance (mitigated ranging error) with different CIR input size dimensions. For each test, we report LoS and NLoS MAE as well as overall standard deviation ($\sigma$). It is clear how a reduced number of input features degrades the performance of the model. Moreover, an input with eight dimensions appears to be the minimum amount of information required to obtain an acceptable range error estimation.}
    \label{fig:CIR reduction}
\end{figure}

In order to select the optimal number of input features, we conduct a grid search study on the number of CIR temporal samples. We progressively reduce the input dimension $K$ from 157, suggested in \cite{bregar2018improving}, to 8. The Mean Absolute Error (MAE) is used as the loss function and metric. Box plot of model performance with the different CIR input sizes is shown in Figure \ref{fig:CIR reduction}.

The network hyperparameters are obtained with an initial random search followed by a grid search exploration to fine-tune them and compromise accuracy and efficiency. We use $N=3$ residual reduction modules with kernel dimensions $k_0=5$ and $k_n=3$, and $F=16$ filters.

Finally, all experimentation adopt Adam as the optimization algorithm \cite{kingma2014adam} with momentum parameters $\beta_{1} = 0.9$, $\beta_{2} = 0.999$, and $\epsilon = 10^{-8}$. The optimal learning rate $\lambda = 3e - 4$ is experimentally derived using the methodology described in \cite{smith2017cyclical} and kept constant for 30 epochs, with a batch size of 32. We employ the TensorFlow framework \footnote{https://www.tensorflow.org/} to train the network on a PC with 32 GB RAM and an Nvidia 2080 Super GP-GPU. The overall training process
can be performed in less than 5 minutes.

\begin{table}[t]
\centering
\resizebox{\columnwidth}{!}{
\begin{tabular}{c!{\quad}c!{\quad}c!{\quad}c!{\quad}c!{\quad}c}
\toprule
Model & CIR & Params & MAE [m] & MAE$_{GO}$ [m] & MAE$_{INT8}$ [m] \\ \hline
\multirow{5}{*}{REMNet} & 157 & 5905  & 0.0687 & 0.0687 & 0.0690 \\
                        & 128 & 5841  & 0.0702 & 0.0702 & 0.0698 \\
                        & 64  & 5713  & 0.0704 & 0.0704 & 0.0701 \\
                        & 32  & 5649  & 0.0710 & 0.0710 & 0.0713 \\
                        & 16  & 5617  & 0.0712 & 0.0712 & 0.0714 \\ \hline
MLP                     & 157 & 54401 & 0.0769 & 0.0777 & 0.0775 \\ \bottomrule
\end{tabular}
}
\small\caption{Model performance for different CIR lengths before and after applying optimizations. The results for a Multilayer Perceptron are included as a reference.}
\label{tab:model_accuracy}
\end{table}

\subsection{Quantitative Results} \label{quantitative_results}

The medium room data samples, used as test set, have a starting MAE of 0.1242 m and a standard deviation of $\sigma=0.1642$ m. The results obtained by the trained reference architectures and their degradation, as optimizations are applied, are shown in Table \ref{tab:model_accuracy}. Each model has been tested five times with different random seeds to obtain statistically significant results. Performances prove the effectiveness of the model, as the MAE of the medium room samples is reduced by $45.7\%$ using the reference model. Consequently, the final error of 0.0687 m is comparable to the actual LoS precision of EVB1000 boards \cite{jimenez2016comparing}. Lastly, REMNet demonstrates to outperform a Multilayer Perceptron (MLP) with around 10\% of its parameters. 

As regards model optimization, columns \textit{CIR} and \textit{Params} report the number of input samples used for the mitigation and the total model parameters, respectively. Moreover, the resulting MAE is reported for all three model configurations: reference, graph optimization (MAE$_{GO}$), and full 8-bit integer quantization (MAE$_{INT8}$). The results show that the effect of graph optimization is null for REMNet, while the MLP performance slightly deteriorates. Integer quantization, instead, minimally increases the resulting MAE for all the models. Finally, our experimentation confirms that fewer dimensions of 128 tend almost linearly to degrade the network's accuracy.

\begin{table}[t]
 \centering
 \resizebox{\columnwidth}{!}{
 \begin{tabular}{c!{\quad}c!{\quad}c!{\quad}c!{\quad}c!{\quad}c}
 \toprule
 Model       & CIR & Params & Dim [kB] & Dim$_{GO}$ [kB] & Dim$_{INT8}$ [kB] \\ \hline
\multirow{5}{*}{REMNet} & 157                  & 5905                & 317.321       & 32.988        & 23.088        \\
     & 128 & 5841  & 317.321 & 32.732  & 23.024 \\
     & 64  & 5713  & 317.157 & 32.220  & 22.896 \\
     & 32  & 5649  & 317.052 & 31.964  & 22.832 \\
     & 16  & 5617  & 317.499 & 31.836  & 22.800 \\ \hline
 MLP & 157 & 54401 & 216.047 & 125.885 & 60.320 \\ \bottomrule
 \end{tabular}
 }
 \small\caption{Model dimensions for different CIR lengths before and after applying optimizations. The results for a Multilayer Perceptron (MLP) are included as a reference.}
 \label{tab:model_dimension}
 \end{table}

\begin{table}[t]
\centering
\resizebox{\columnwidth}{!}{
\begin{tabular}{c!{\quad}c!{\quad}c!{\quad}c!{\quad}c!{\quad}c!{\quad}c}
\toprule
Model & CIR & $f_{m}$ [Hz] & $V_{cc}$ [V] & $I_{abs}$ [mA] & $P_{abs}$ [mW] & $E_{inf}$ [mJ] \\ \hline
\multirow{5}{*}{REMNet} & 157 & 17.2  & 3.3 & 16.2 & 53.4 & 3.1  \\
                                 & 128 & 21.2  & 3.3 & 16.0 & 52.8 & 2.5  \\
                                 & 64  & 41.0  & 3.3 & 15.8 & 52.2 & 1.3  \\
                                 & 32  & 77.8  & 3.3 & 15.6 & 51.6 & 0.66 \\
                                 & 16  & 140.0 & 3.3 & 15.6 & 51.6 & 0.37 \\ \hline
MLP                     & 157 & 184.1 & 3.3 & 16.2 & 53.4 & 0.29 \\ 
\hline
\end{tabular}
}
\small\caption{Real-time performance for different optimized models, including inference frequency, consumed power, and network energy usage.}
\label{tab:model_frequency}
\end{table}

Despite the insignificant effect of graph optimization on performance, memory occupancy greatly shrinks. Our results show a reduction of about 90\% for REMNet, while the MLP only halves its memory footprint due to its higher number of parameters. In addition, quantization allows a further reduction of REMNet memory requirements of an additional 30\%, confirming the great benefit of using both optimization and quantization techniques provided by TensorFlow Lite converter. The MLP reduces another 50\% of its memory footprint with full integer quantization, reaching a final size of about three times REMNet. Therefore, the proposed model can outperform the baseline both in error mitigation capability and memory requirements. All the results on the memory footprint of the models under examination are presented in Table \ref{tab:model_dimension}.

Finally, the inference speed and the power consumed by the Arduino board for each considered model are analyzed and presented in Table \ref{tab:model_frequency}. The frequencies, denoted as {f$_{m}$}, have been measured as the reciprocal of the maximum inference time over a series of tests. In all the cases, they can be considered compliant for real-time applications. In particular, the MLP requires less computational effort despite the more significant number of parameters because it involves simpler math operations than REMNet. Moreover, reducing CIR length results in an almost linear increase in inference speed. To assess power consumption, instead, we measured the absorbed current $I_{abs}$ with a voltage supply $V_{cc} = 3.3$ V. Results show that, as the microcontroller is constantly processing data, the power usage can be considered constant for all the cases. However, since inference speed significantly changes with model complexity, we computed the energy required for a single inference step $E_{inf}$ by dividing the consumed power $P_{abs}$ by the frequency $f_{m}$. It is noticeable that very few mJ are sufficient to run range mitigation, reaching values under 1 mJ. That proves that the already efficient design of the proposed model, in conjunction with 8-bit weight precision and graph optimization techniques, makes deep learning a feasible solution for effective ultra-low-power UWB range error mitigation.


\section{Conclusions} \label{conclusions}
In this paper, we introduced UWB range error mitigation for ultra-low-power microcontrollers. Our effective and power-efficient methodology builds on top of the latest advancement in deep learning and neural networks optimization techniques to provide precise localization in NLOS and LOS conditions with a few mW of power. We proposed a modified version of REMNet, a lightweight model designed explicitly for range error mitigation on ultra-low-power AI edge devices.
Our extensive experimentation proves how the proposed system successfully runs at a high frequency on microcontrollers and provides enhanced localization in indoor environments. Future works will integrate the proposed methodology on a compact custom board designed around the DWM1001C DecaWave module to provide a compact solution for precise localization.

\section*{Acknowledgments}
This work is partially supported by the Italian government via the NG-UWB project (MIUR PRIN 2017) and developed with the contribution of the Politecnico di Torino Interdepartmental Center for Service Robotics PIC4SeR\footnote{https://pic4ser.polito.it} and SmartData@Polito\footnote{https://smartdata.polito.it}.

%
%
\footnotesize
\bibliography{main.bib}
\bibliographystyle{abbrv} 
\end{document}